\documentclass[10pt,twocolumn,letterpaper]{article}

\usepackage{cvpr}
\usepackage{times}
\usepackage{graphicx}
\usepackage{amsmath}
\usepackage{amssymb}
\usepackage{booktabs}
\usepackage{multirow}
\usepackage[table]{xcolor}   
\usepackage{subcaption}         
\usepackage{bm}                 
\usepackage{tikz}               
\usetikzlibrary{arrows.meta,calc,positioning,shapes.geometric}
\usepackage[pagebackref=true,breaklinks=true,colorlinks,bookmarks=false]{hyperref}

\renewcommand{\etal}{\textit{et~al.}\xspace}

\newcommand{\metric}[1]{\textsc{#1}}
\newcommand{\nids}{n_{\text{ids}}}

\newcommand{\cmark}{\ding{51}}
\newcommand{\xmark}{\ding{55}}
\usepackage{pifont}    

\newcommand\blfootnote[1]{%
  \begingroup
  \renewcommand\thefootnote{}\footnote{#1}%
  \addtocounter{footnote}{-1}%
  \endgroup
}

\AtBeginDocument{%
  \setlength{\abovedisplayskip}{5pt plus 2pt minus 2pt}%
  \setlength{\belowdisplayskip}{5pt plus 2pt minus 2pt}%
  \setlength{\abovedisplayshortskip}{2pt plus 1pt}%
  \setlength{\belowdisplayshortskip}{4pt plus 2pt minus 2pt}%
}

\begin{document}

\title{Quantifying Training Membership Information in the\\ Hyperspherical Embedding Geometry of Face Recognition Models}
\author{\vspace{5pt}{\"U}nsal {\"O}zt{\"u}rk$^{1}$\hspace{1.5em}S{\'e}bastien Marcel$^{1,2}$\\
	$^{1}$Idiap Research Institute, Martigny, Switzerland\\
	\vspace{3pt}$^{2}$Universit\'{e} de Lausanne (UNIL), Lausanne, Switzerland\\
\texorpdfstring{{\small\texttt{\{unsal.ozturk, sebastien.marcel\}@idiap.ch}}}{{unsal.ozturk, sebastien.marcel}@idiap.ch}\\
}
\maketitle
\thispagestyle{empty}
\blfootnote{This work has been submitted to the IEEE for possible publication. Copyright may be transferred without notice, after which this version may no longer be accessible.}

\begin{abstract}
Face recognition models represent each face as an embedding vector on the unit hypersphere by clustering embeddings of the same identity while pushing different identities apart through angular-margin losses. Because these losses act only on training identities, non-member identities may form clusters with different geometric properties. In this paper, we quantify the magnitude of this difference and what training-time factors control it. We compute four statistics based on cluster geometry across 180 face recognition models in a factorial design over IResNet backbone size, loss head, training duration, and the number of training identities, and evaluate each configuration on nine benchmarks. Our results indicate that the number of training identities has the largest effect on member/non-member separability, while backbone and loss head contribute far less, and that, on a same-domain held-out reference, the geometric membership signal decreases monotonically as more identities are added to training. We provide an analysis of cross-domain (pose, age, quality, ethnicity) non-member benchmarks and report that these inflate the apparent membership signal. Finally, we fuse all four statistics with a learned classifier to reveal additional membership information beyond the best individual statistic.
\end{abstract}
\vspace{-10pt} 
\section{Introduction}
\label{sec:intro}

%
%
\begin{figure}[t]
\centering
\begin{tikzpicture}[
  >=Stealth, font=\footnotesize,
  box/.style  = {draw, rounded corners=2pt, align=center},
  mbox/.style = {box, fill=gray!12, minimum width=2.8cm,
                 minimum height=8mm, font=\footnotesize\bfseries},
  cbox/.style = {box, fill=yellow!8, minimum height=6.5mm,
                 inner xsep=5pt, font=\footnotesize\itshape},
  dbox/.style = {box, minimum height=6.5mm,
                 font=\footnotesize\bfseries, inner xsep=5pt,
                 inner ysep=2pt},
  arr/.style  = {->, thick, gray!55!black},
  darr/.style = {->, thick},
]

\node[inner sep=0] (personM) at (-1.20, 0) {};
\begin{scope}[shift={(-1.20, 0)}]
  \fill[red!45] (0,0.24) circle(0.16);
  \fill[red!45, rounded corners=1pt]
    (-0.24,-0.09) .. controls (-0.24,0.04) and (-0.13,0.08) ..
    (0,0.08) .. controls (0.13,0.08) and (0.24,0.04) ..
    (0.24,-0.09) -- cycle;
  \node[font=\scriptsize\bfseries, text=red!70!black] at (0,-0.22) {Member Identities};
\end{scope}
\node[inner sep=0] (personN) at (1.20, 0) {};
\begin{scope}[shift={(1.20, 0)}]
  \fill[blue!45] (0,0.24) circle(0.16);
  \fill[blue!45, rounded corners=1pt]
    (-0.24,-0.09) .. controls (-0.24,0.04) and (-0.13,0.08) ..
    (0,0.08) .. controls (0.13,0.08) and (0.24,0.04) ..
    (0.24,-0.09) -- cycle;
  \node[font=\scriptsize\bfseries, text=blue!70!black] at (0,-0.22) {Non-member Identities};
\end{scope}

\draw[arr] (-1.20,-0.50) -- (-1.20,-1.20) -- (-0.40,-1.20) -- (-0.40,-1.76);
\draw[arr] ( 1.20,-0.50) -- ( 1.20,-1.20) -- ( 0.40,-1.20) -- ( 0.40,-1.76);

\node[mbox] (model) at (0,-2.20) {Black-box FR $f(\cdot)$};

\node[font=\scriptsize, text=red!60!black]  at (-2.75,-1.62) {Member embeddings};
\node[font=\scriptsize, text=blue!60!black] at ( 2.75,-1.62) {Non-member embeddings};

\begin{scope}[shift={(-3.15,-1.97)}]
  \fill[red!70] (+0.072,+0.050) circle(0.033);
  \fill[red!68] (-0.001,-0.055) circle(0.033);
  \fill[red!66] (-0.163,+0.052) circle(0.033);
  \fill[red!64] (+0.169,+0.023) circle(0.033);
  \fill[red!62] (+0.120,-0.042) circle(0.033);
  \fill[red!60] (+0.023,-0.072) circle(0.033);
  \fill[red!58] (-0.068,+0.035) circle(0.033);
  \draw[red!50, semithick, rotate=0] (0,0) ellipse(0.27 and 0.10);
\end{scope}

\begin{scope}[shift={(-2.35,-1.97)}]
  \fill[red!70] (-0.054,-0.002) circle(0.033);
  \fill[red!68] (-0.012,-0.043) circle(0.033);
  \fill[red!66] (-0.092,-0.166) circle(0.033);
  \fill[red!64] (+0.036,-0.122) circle(0.033);
  \fill[red!62] (-0.016,+0.160) circle(0.033);
  \fill[red!60] (+0.008,-0.095) circle(0.033);
  \fill[red!58] (+0.020,+0.119) circle(0.033);
  \draw[red!50, semithick, rotate=70] (0,0) ellipse(0.28 and 0.11);
\end{scope}

\begin{scope}[shift={(-3.15,-2.43)}]
  \fill[red!70] (-0.064,+0.175) circle(0.033);
  \fill[red!68] (-0.065,+0.105) circle(0.033);
  \fill[red!66] (-0.047,+0.104) circle(0.033);
  \fill[red!64] (+0.013,-0.024) circle(0.033);
  \fill[red!62] (+0.024,-0.153) circle(0.033);
  \fill[red!60] (-0.029,+0.101) circle(0.033);
  \fill[red!58] (+0.043,+0.049) circle(0.033);
  \draw[red!50, semithick, rotate=-55] (0,0) ellipse(0.26 and 0.10);
\end{scope}

\begin{scope}[shift={(-2.35,-2.43)}]
  \fill[red!70] (+0.168,+0.031) circle(0.033);
  \fill[red!68] (+0.024,+0.084) circle(0.033);
  \fill[red!66] (+0.047,+0.083) circle(0.033);
  \fill[red!64] (-0.084,-0.003) circle(0.033);
  \fill[red!62] (-0.106,-0.100) circle(0.033);
  \fill[red!60] (+0.024,+0.109) circle(0.033);
  \fill[red!58] (+0.078,+0.047) circle(0.033);
  \draw[red!50, semithick, rotate=30] (0,0) ellipse(0.29 and 0.11);
\end{scope}

\begin{scope}[shift={(2.35,-1.97)}]
  \fill[blue!70] (-0.006,+0.144) circle(0.033);
  \fill[blue!68] (-0.083,-0.041) circle(0.033);
  \fill[blue!66] (+0.153,-0.025) circle(0.033);
  \fill[blue!64] (-0.185,-0.026) circle(0.033);
  \fill[blue!62] (-0.132,-0.167) circle(0.033);
  \fill[blue!60] (+0.092,-0.066) circle(0.033);
  \fill[blue!58] (+0.179,+0.101) circle(0.033);
  \draw[blue!50, semithick, rotate=30] (0,0) ellipse(0.38 and 0.16);
\end{scope}

\begin{scope}[shift={(3.15,-1.97)}]
  \fill[blue!70] (-0.146,+0.082) circle(0.033);
  \fill[blue!68] (-0.146,+0.055) circle(0.033);
  \fill[blue!66] (+0.018,+0.122) circle(0.033);
  \fill[blue!64] (+0.197,-0.075) circle(0.033);
  \fill[blue!62] (-0.059,+0.084) circle(0.033);
  \fill[blue!60] (-0.160,+0.150) circle(0.033);
  \fill[blue!58] (-0.006,-0.114) circle(0.033);
  \draw[blue!50, semithick, rotate=-10] (0,0) ellipse(0.40 and 0.17);
\end{scope}

\begin{scope}[shift={(2.35,-2.43)}]
  \fill[blue!70] (-0.142,-0.178) circle(0.033);
  \fill[blue!68] (-0.128,+0.026) circle(0.033);
  \fill[blue!66] (-0.125,-0.230) circle(0.033);
  \fill[blue!64] (-0.162,-0.098) circle(0.033);
  \fill[blue!62] (-0.217,-0.103) circle(0.033);
  \fill[blue!60] (+0.097,+0.238) circle(0.033);
  \fill[blue!58] (+0.151,+0.079) circle(0.033);
  \draw[blue!50, semithick, rotate=50] (0,0) ellipse(0.39 and 0.16);
\end{scope}

\begin{scope}[shift={(3.15,-2.43)}]
  \fill[blue!70] (+0.170,-0.151) circle(0.033);
  \fill[blue!68] (-0.184,+0.147) circle(0.033);
  \fill[blue!66] (-0.261,+0.086) circle(0.033);
  \fill[blue!64] (+0.077,-0.065) circle(0.033);
  \fill[blue!62] (+0.261,-0.188) circle(0.033);
  \fill[blue!60] (-0.179,+0.013) circle(0.033);
  \fill[blue!58] (+0.256,-0.077) circle(0.033);
  \draw[blue!50, semithick, rotate=-25] (0,0) ellipse(0.41 and 0.15);
\end{scope}

\draw[arr] (model.west) -- (-2.07,-2.20);
\draw[arr] (model.east) -- ( 1.97,-2.20);

\draw[arr] (-2.75,-2.60) -- (-2.75,-3.00);
\draw[arr] ( 2.75,-2.60) -- ( 2.75,-3.00);
\node[font=\scriptsize\itshape, text=gray!60!black] at (0,-2.80)
  {extract per-cluster statistics};


\draw[rounded corners=3pt, gray!50, line width=0.4pt]
  (-3.45,-3.00) rectangle (-0.10,-5.30);
\node[font=\scriptsize\bfseries, anchor=north west, text=gray!60!black]
  at (-3.33,-3.04) {Case 1: Separated};
\begin{scope}[shift={(-2.55,-4.20)}]
  \draw[gray!30, thin] (-0.50,0) -- (0.50,0);
  \draw[blue!60, semithick, smooth, samples=30, domain=-0.48:0.48]
    plot (\x, {0.50*exp(-(\x+0.16)*(\x+0.16)/0.009)});
  \draw[red!70, semithick, smooth, samples=30, domain=-0.48:0.48]
    plot (\x, {0.55*exp(-(\x-0.17)*(\x-0.17)/0.008)});
  \draw[black, semithick] (0.04,0) -- (0.04,0.55);
  \node[font=\tiny, anchor=south west] at (0.07,0.42) {$\tau$};
  \node[font=\scriptsize, text=gray!70] at (0,-0.22) {$s_1$};
\end{scope}

\begin{scope}[shift={(-0.90,-4.20)}]
  \draw[gray!30, thin] (-0.50,0) -- (0.50,0);
  \draw[red!70, semithick, smooth, samples=30, domain=-0.48:0.48]
    plot (\x, {0.36*exp(-(\x+0.14)*(\x+0.14)/0.020)});
  \draw[blue!60, semithick, smooth, samples=30, domain=-0.48:0.48]
    plot (\x, {0.30*exp(-(\x-0.13)*(\x-0.13)/0.024)});
  \draw[black, semithick] (-0.04,0) -- (-0.04,0.55);
  \node[font=\tiny, anchor=south east] at (-0.07,0.42) {$\tau$};
  \node[font=\scriptsize, text=gray!70] at (0,-0.22) {$s_2$};
\end{scope}
\draw[red!70,  semithick] (-3.30,-4.62) -- (-3.02,-4.62)
  node[right, font=\scriptsize, text=red!60!black] {member};
\draw[blue!60, semithick] (-3.30,-4.82) -- (-3.02,-4.82)
  node[right, font=\scriptsize, text=blue!60!black] {non-member};

\node[font=\scriptsize\itshape, text=gray!70!black, text width=2.8cm, align=center]
  at (-1.775,-5.06) {$\tau$ separates distributions};

\draw[rounded corners=3pt, gray!50, line width=0.4pt]
  (0.10,-3.00) rectangle (3.45,-5.30);
\node[font=\scriptsize\bfseries, anchor=north east, text=gray!60!black]
  at (3.33,-3.04) {Case 2: Overlapping};
\begin{scope}[shift={(0.90,-4.20)}]
  \draw[gray!30, thin] (-0.50,0) -- (0.50,0);
  \draw[blue!60, semithick, smooth, samples=30, domain=-0.48:0.48]
    plot (\x, {0.46*exp(-(\x+0.02)*(\x+0.02)/0.018)});
  \draw[red!70, semithick, smooth, samples=30, domain=-0.48:0.48]
    plot (\x, {0.48*exp(-(\x-0.02)*(\x-0.02)/0.016)});
  \node[font=\scriptsize, text=gray!70] at (0,-0.22) {$s_1$};
\end{scope}

\begin{scope}[shift={(2.55,-4.20)}]
  \draw[gray!30, thin] (-0.50,0) -- (0.50,0);
  \draw[red!70, semithick, smooth, samples=30, domain=-0.48:0.48]
    plot (\x, {0.30*exp(-(\x+0.01)*(\x+0.01)/0.035)});
  \draw[blue!60, semithick, smooth, samples=30, domain=-0.48:0.48]
    plot (\x, {0.28*exp(-(\x-0.01)*(\x-0.01)/0.040)});
  \node[font=\scriptsize, text=gray!70] at (0,-0.22) {$s_2$};
\end{scope}
\draw[red!70,  semithick] (0.20,-4.62) -- (0.48,-4.62)
  node[right, font=\scriptsize, text=red!60!black] {member};
\draw[blue!60, semithick] (0.20,-4.82) -- (0.48,-4.82)
  node[right, font=\scriptsize, text=blue!60!black] {non-member};

\node[font=\scriptsize\itshape, text=gray!70!black, text width=2.8cm, align=center]
  at (1.775,-5.06) {thresholding not possible};

\node[dbox, fill=red!12, text=red!80!black]
  (leak)   at (-1.775,-6.20) {Membership signal};
\node[dbox, fill=blue!10, text=blue!70!black]
  (noleak) at ( 1.775,-6.20) {No Membership signal};

\draw[arr] (-1.775,-5.30) -- (leak.north);
\draw[arr] ( 1.775,-5.30) -- (noleak.north);

\end{tikzpicture}
\caption{Measurement pipeline. Per-identity statistics ($s_1, s_2, \ldots$) are computed from the embedding clusters of a face recognition model. If the distributions of these statistics differ between member identities (present in the training set) and non-member identities (absent from it) as shown on the left, the explicit cluster geometry carries membership information on these statistics; if they coincide (right), it does not.}
\label{fig:mia-pipeline}
\vspace{-14pt} 
\end{figure}
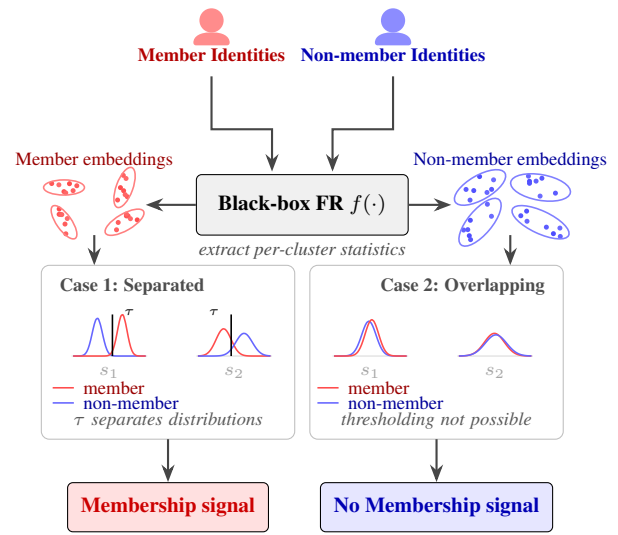

Modern face recognition (FR) systems are trained on large-scale datasets of face crops, typically collected from the internet~\cite{Zhu2021WebFace260M}. To represent a person, the system takes multiple crops of that person's face and passes each through a deep neural network that produces a 512-dimensional embedding vector, normalised to lie on the unit hypersphere. The training of these models is primarily driven by angular-margin losses~\cite{deng2019arcface,wang2018cosface,meng2021magface}, which pull the embeddings of crops belonging to the same identity into a cluster on the hypersphere while pushing apart the clusters of different identities. During training, a classification head maps embeddings to identity logits and this head is discarded after training. When deployed, the system typically uses only the backbone neural network. Enrolled faces are stored as embeddings, and the similarity between two faces is measured by the inner product of their corresponding embeddings.

As angular-margin training directly optimises the compactness of each identity's cluster, we pose the following question: how much information about training-set membership is retained in the geometry of these embedding clusters? If training identities form systematically tighter or better-separated clusters than unseen identities, then the embedding geometry carries a residual membership signal.

Quantifying this residual signal is related to membership inference attacks (MIA), a line of work investigating whether a given sample or identity was part of a model's training data~\cite{Shokri2017MIA,Carlini2022MIAFirstPrinciples}. In practice, membership inference serves as an auditing tool for data privacy by allowing individuals or regulators to verify whether a model was trained on data belonging to a specific person in the context of data protection rights such as the right to erasure under the GDPR~\cite{GDPR2016}. MIA literature primarily focuses on closed-set classifiers that expose prediction confidences, loss values, or the model weights directly~\cite{Shokri2017MIA,Salem2019MLLeaks, Carlini2022MIAFirstPrinciples,Hu2022MIASurvey}. In our setting, we consider open-set face recognition where neither logits nor losses nor model weights are available, so the membership signal must be sought entirely in the geometry of the embeddings themselves.

To measure this signal, we compute four different statistics capturing different aspects of cluster geometry on the hypersphere and measure how much they differ between training and unseen identities (Figure~\ref{fig:mia-pipeline}). We train 180 face recognition models from scratch (Section~\ref{sec:ablation}) and evaluate each model on nine benchmarks, including a same-domain held-out split that controls for the pose, age, quality, and ethnicity shifts present in the other benchmarks. To our knowledge, this constitutes the largest systematic study of membership information in face embedding geometry. Our contributions are: \textbf{(1)}~we quantify the membership information retained in four geometric statistics computed from face embeddings; \textbf{(2)}~we identify the primary factors that control how much of this information persists; \textbf{(3)}~we show that cross-domain non-member benchmarks (e.g.\ differing in pose or age) inflate the apparent membership signal by conflating domain shift with training-attributable differences; \textbf{(4)}~we demonstrate that fusing all statistics captures membership information beyond the best individual statistic.

\section{Related Work}
\label{sec:related}

\textbf{Membership inference on classifiers.} 
Membership inference literature targets closed-set classifiers, where the attacker can observe prediction confidences, logits, or loss values of the target model. In the open-set face recognition setting, however, none of these signals are available and this has motivated a line of work operating directly on embeddings. Shokri~\etal~\cite{Shokri2017MIA} introduce the shadow-model paradigm and train substitute classifiers to learn a binary meta-classifier on output distributions. Salem~\etal~\cite{Salem2019MLLeaks} relax many of these assumptions and show that a single shadow model and only the top-$k$ predictions often suffice. Carlini~\etal~\cite{Carlini2022MIAFirstPrinciples} recast MIA as a likelihood-ratio attack (LiRA) and derive near-optimal attacks. They show that per-example hardness varies by orders of magnitude. Hu~\etal~\cite{Hu2022MIASurvey} provide a comprehensive survey. 

\begin{table}[!t]
\centering
\caption{%
  \textbf{Comparison with prior membership inference studies.}
  Level: identity-level or sample-level membership question.
  \#\,Mod.: number of target models evaluated.
  Fact.: whether a factorial experimental design was used.
}
\label{tab:related-summary}
\setlength{\tabcolsep}{3pt}
\renewcommand{\arraystretch}{1.10}
\footnotesize
\begin{tabular}{@{} l l c c c @{}}
  \toprule
  \textbf{Work}
    & \textbf{Domain}
    & \textbf{Level}
    & \textbf{\#\,Mod.}
    & \textbf{Fact.} \\
  \midrule
  \multicolumn{5}{@{}l}{\textit{Embedding-level (black-box)}} \\[2pt]
  Li~\etal~\cite{Li2022UserLevelMIA}
    & FR / re-identification & Identity & 6 & \xmark \\
  FACE-AUD.~\cite{Chen2023FaceAuditor}
    & FR / few-shot & Identity & 4 & \xmark \\
  SLMIA-SR~\cite{Chen2024SLMIASR}
    & Speaker recognition & Identity & 5 & \xmark \\
  EncoderMI~\cite{Liu2021EncoderMI}
    & Contrastive learning & Sample & 4 & \xmark \\
  \midrule
  \multicolumn{5}{@{}l}{\textit{Model-internals (white-box)}} \\[2pt]
  MINT~\cite{DeAlcala2025MINT}
    & FR & Sample & 3 & \xmark \\
  DeAlcala~\cite{DeAlcala2024Factors}
    & FR & Sample & 9 & \xmark$^\ast$ \\
  Mancera~\cite{Mancera2025MINTObject}
    & Object classification & Sample & 4 & \xmark \\
  Huang~\cite{Huang2024InferenceNoClassHead}
    & FR & Sample & 2 & \xmark \\
  \midrule
  \textbf{Ours}
    & \textbf{FR} & \textbf{Identity}
    & \textbf{180} & \cmark \\
  \bottomrule
\end{tabular}\\[1pt]
{\scriptsize $^\ast$Varies factors individually but not in a fully crossed design.}

\vspace{-18pt} 

\end{table}

\textbf{Membership inference on embedding models.}
Existing embedding-based approaches differ in the granularity of the membership question: some ask whether a specific image was used for training (sample level), while others ask whether \emph{any} image of a given person was included (identity level). Li~\etal~\cite{Li2022UserLevelMIA} formalise identity-level inference on metric-embedding models and use intra-similarity and inter-dissimilarity of probe embeddings as attack features on person re-identification and face recognition. Chen~\etal~\cite{Chen2023FaceAuditor} propose FACE-AUDITOR, a system that queries few-shot face recognition models with designed probing sets and uses the returned similarity scores as auditing features. They report accuracies up to 99\%. SLMIA-SR~\cite{Chen2024SLMIASR} addresses the analogous problem in speaker recognition by training shadow models with controlled membership mixing ratios. All three operate at the identity level. EncoderMI~\cite{Liu2021EncoderMI}, by contrast, operates at the sample level on contrastive-learning encoders including CLIP and exploits the observation that an overfitted encoder produces more similar representations for augmented views of a training sample than for an unseen one.

\textbf{White-box approaches on face recognition.}
A parallel line of work allows access to model internals. The motivation is that intermediate representations such as activation maps, batch-normalisation statistics, or gradient signals may carry a stronger membership trace than the final embedding alone, at the cost of requiring more privileged access to the model. These methods also typically operate at the sample level, determining whether a specific image (rather than a person) was included in training. The Membership Inference Test (MINT) of DeAlcala~\etal~\cite{DeAlcala2025MINT} trains an MLP or CNN on intermediate activation maps to classify individual face images as member or non-member and achieves up to 90\% accuracy on three ResNet-100 models. A companion study~\cite{DeAlcala2024Factors} analyses factors that affect MINT performance, including loss function, dropout, and number of training epochs. Mancera~\etal~\cite{Mancera2025MINTObject} extend MINT to object classification and report 70--80\% precision depending on layer depth. Huang~\etal~\cite{Huang2024InferenceNoClassHead} exploit distances between intermediate features and batch-normalisation parameters for sample-level membership and model-inversion attacks~\cite{Fredrikson2015Inversion}. Even with this level of access, Rezaei and Liu~\cite{Rezaei2021DifficultyMIA} show that the advantage over random guessing can be modest once the model generalises well.

\textbf{Memorisation and embedding geometry.}
The question of why a geometric membership signal might exist at all connects to a discussion on memorisation in deep networks. If a model must memorise certain training examples to achieve low error, particularly on the tail of the data distribution, then its internal representations and output embeddings may carry traces of those examples. Feldman~\cite{Feldman2020Memorization} formalises this argument, showing theoretically that long-tailed distributions necessitate memorisation for near-optimal generalisation. Angular-margin losses~\cite{deng2019arcface,wang2018cosface, meng2021magface} amplify this effect, potentially making training identities geometrically distinguishable from unseen ones. Unlearning methods~\cite{Bourtoule2021SISA, Golatkar2020Forgetting} and differential privacy~\cite{Abadi2016DP} aim to erase or bound such signals.

\textbf{Comparison with this work.}
Our contribution is a controlled measurement study of how much membership information geometric statistics carry, not a stronger attack feature. Table~\ref{tab:related-summary} summarises the literature. Prior embedding-level work proposes specific attack features and demonstrates that membership inference is feasible through evaluating a small number of models trained under a single configuration. Our study differs in three respects. First, the goal: we measure how much membership information is retained in the explicit cluster geometry and identify which training factors control the amount that persists. Second, the experimental scale: 180 models, whereas the largest prior study evaluates nine. Third, domain control: we add a same-domain held-out reference, which aims to separate training-attributable signal from domain-shift effects. Several of the per-identity statistics we evaluate have already been used as features in prior attacks. For instance, the intra-similarity score proposed by Li~\etal~\cite{Li2022UserLevelMIA} is equivalent to the mean pairwise cosine similarity that we include as a baseline statistic (defined in Section~\ref{sec:metrics}).

\section{Methodology}
\label{sec:method}

\subsection{Access Model and Measurement Procedure}
\label{sec:threat}

We assume that an auditor can query an FR model~$f$ with $k \geq 2$ probe images of a target identity~$y$ and obtain the corresponding $\ell_2$-normalised embeddings. From the resulting cluster, the auditor computes four per-identity statistics (Section~\ref{sec:metrics}) and uses them to assess if the cluster geometry carries information about $y$'s membership in the training set. For each model we train, we compute these statistics on member identities (from WebFace4M) and on non-member identities drawn from the benchmarks in Table~\ref{tab:datasets}.

%

\begin{table*}[!t]
\centering
\caption{%
  \textbf{Dataset overview for membership evaluation.}
  ``Eval.\ subj.'': identities with ${\geq}2$ crops (required for
  per-identity cluster statistics).
  \textsuperscript{$\dagger$}For WebFace4M this is the held-out split we sample
  for evaluation; the dataset itself contains far more identities with
  ${\geq}2$ crops.
}
\label{tab:datasets}
\setlength{\tabcolsep}{3pt}
\footnotesize
\begin{tabular}{@{}llrrrl@{}}
\toprule
Dataset & Primary variation
  & Subjects & Eval.\ subj. & Images
  & Covariate metadata \\
\midrule
\rowcolor{red!6}
\multicolumn{6}{l}{\textit{Cross-distribution benchmarks (non-member identities)}} \\[2pt]
LFW~\cite{LFWTech}
  & Unconstrained (baseline)
  & 5{,}749 & 1{,}680 & 13{,}233
  & \textemdash{} (subject name only) \\
CPLFW~\cite{zheng2018cross}
  & Cross-pose (subset of LFW identities)
  & 3{,}929 & 3{,}909 & 11{,}652
  & yaw (estimated from landmarks) \\
CFP-FF~\cite{Sengupta2016CFP}
  & Frontal-only probes
  & \multirow{2}{*}{500} & 500 & 7{,}000
  & pose $\in$ \{frontal\} \\
CFP-FP~\cite{Sengupta2016CFP}
  & Frontal+profile probes
  &  & 500 & 7{,}000
  & pose $\in$ \{frontal, profile\} \\
AgeDB-30~\cite{Moschoglou2017AgeDB}
  & Large intra-identity age gap ($\geq$30\,yr)
  & 440 & 440 & 12{,}240
  & age (3--100\,yr), gender \\
XQLFW~\cite{Knoche2021XQLFW}
  & Low image quality (paired with LFW)
  & 5{,}749 & 1{,}672 & 13{,}233
  & \textemdash{} (quality implicit by design) \\
RFW~\cite{Wang2019RFW}
  & Ethnicity-balanced (4 groups)
  & 11{,}429 & 11{,}429 & 40{,}607
  & ethnicity, gender, nationality \\
\midrule
\rowcolor{blue!6}
\multicolumn{6}{l}{\textit{Large-scale benchmarks}} \\[2pt]
IJB-C~\cite{Maze2018IJBC}
  & Unconstrained, mixed image+video
  & 3{,}531 & 3{,}531 & 469{,}376
  & ethnicity, gender, nationality \\
\midrule
\rowcolor{green!6}
\multicolumn{6}{l}{\textit{Training data and within-distribution evaluation}} \\[2pt]
WebFace4M~\cite{Zhu2021WebFace260M}
  & Large-scale training source
  & 205{,}990 & ${\leq}\,5{,}000$\textsuperscript{$\dagger$} & 4{,}000{,}000
  & \textemdash{} (subject ID only) \\
\bottomrule
\end{tabular}
\vspace{-10pt} 
\end{table*}

For each statistic, we obtain a scalar score per identity. To quantify how much membership information a statistic carries, we sweep a decision threshold~$\tau$ across the range of observed values and classify an identity as a member if its score exceeds~$\tau$ (or falls below~$\tau$ for statistics where lower values indicate membership). We report the area under the resulting ROC curve (AUC), hereafter threshold-classifier AUC, as a threshold-free summary, which is equal to the probability that a randomly chosen member scores higher than a randomly chosen non-member. This serves as our primary measure of the membership information carried by a given statistic. We use AUC rather than the true positive rate at a fixed false positive rate, because our framing concerns separability across all decision thresholds rather than the operating point most relevant for a deployable attack. Of the four statistics defined in the next section, two require only the embeddings and two additionally require the training-time scale and margin hyperparameters. We refer to these two access levels as black-box (embeddings only) and grey-box (embeddings plus $s$ and $m$). The grey-box setting is realistic when models are released alongside their training configurations.

\subsection{Per-Identity Geometric Statistics}
\label{sec:metrics}

Given an identity $y$ with $k \geq 2$ probe images, we extract embeddings $\{e_1, \ldots, e_k\}$ via the target model and compute four statistics from the resulting cluster. Each statistic captures a different aspect of cluster shape. The underlying intuition is that an angular-margin loss explicitly optimises the representation of training identities, so members are expected to form tighter clusters on the embedding hypersphere than non-members. If that expectation holds, the statistic will carry membership information; the question is how much, and what factors control it.

\textbf{Pairwise cosine similarity (\metric{PairCos}).}
The mean cosine similarity over pairs of embeddings within the
identity:
\begin{equation*}
    \bar{s} = \binom{k}{2}^{-1} \sum_{i < j} \cos(e_i, e_j).
\end{equation*}
Higher values indicate a more coherent cluster.

\textbf{vMF concentration $\kappa$.}
We model the per-identity embedding distribution as a von Mises-Fisher
(vMF) distribution on the unit hypersphere and estimate the
concentration parameter $\kappa$ via maximum likelihood.
We use the closed-form approximation of
Banerjee~\etal~\cite{banerjee2005vmf}:
\begin{equation*}
    \hat{\kappa}_0 = \frac{\bar{R}\,(d - \bar{R}^2)}{1 - \bar{R}^2},
    \qquad
    \bar{R} = \left\| \frac{1}{k}\sum_{i=1}^{k} e_i \right\|,
\end{equation*}
where $d = 512$ is the embedding dimensionality and $\bar{R}$ is the
mean resultant length.
Higher $\hat{\kappa}$ corresponds to greater directional concentration.
Note that ordering identities by
$\hat{\kappa}$ is equivalent to ordering them by cluster inertia
$I = k^{-1}\sum_i \|e_i - \mu\|^2$, where $\mu = k^{-1}\sum_i e_i$
is the cluster mean.
For $\ell_2$-normalised embeddings, $I$ reduces to
$1 - \bar{R}^2$, and $\hat{\kappa}_0$ is
strictly decreasing in $1-\bar{R}^2$, so the two
quantities produce identical identity rankings.

\textbf{Penalised logit (\metric{PenLogit}).}
To approximate the margin-penalised logit that the training loss
would compute for the cluster, we use a leave-one-out scheme.
For each embedding $e_i$, we compute a prototype
$\hat{w}_{\backslash i} = \bar{e}_{\backslash i}/\|\bar{e}_{\backslash i}\|$
as the $\ell_2$-normalised mean of the remaining $k{-}1$
embeddings, and the margin-penalised logit is
\begin{equation*}
    \ell_i = s \cdot g\left(\cos\angle\left(e_i, \hat{w}_{\backslash i}\right),\; m\right),
\end{equation*}
where $s$ and $m$ are the scale and margin of the loss head, and $g$
is the head-specific penalisation function (e.g.\
$g(\cos\theta, m) = \cos(\theta + m)$ for ArcFace).
The statistic value is the mean $\bar{\ell} = k^{-1}\sum_i \ell_i$.
A higher penalised logit indicates that the embeddings are tightly
aligned to their own prototype.

\textbf{Prototype softmax CE (\metric{ProtoCE}).}
We construct a surrogate of the training loss.
For each identity $y$, we compute a class prototype
$\hat{w}_y = \mu / \|\mu\|$ and treat all other identities in
the evaluation set as negative classes.
The cross-entropy loss with the margin-modified logit for the target
class and standard cosine logits for the negatives gives
\begin{equation*}
    \mathcal{L}_y =
        -\log \frac{\exp\bigl(s\cdot g(\cos\theta_y,\, m)\bigr)}
             {\exp\bigl(s\cdot g(\cos\theta_y,\, m)\bigr) +
              \sum_{j \neq y} \exp(s\cdot\cos\theta_j)}
\end{equation*}
We report $-\mathcal{L}_y$ so that higher values indicate membership,
consistent with the sign convention of the other statistics.
Because the softmax denominator depends on the composition and size
of the evaluation set, \metric{ProtoCE} values are not directly
comparable across benchmarks with different identity pools.
Throughout, all AUC values are oriented so that $\text{AUC} > 0.5$
indicates better-than-chance separation.

\section{Experimental Setup}
\label{sec:setup}

\subsection{Training Data}
\label{sec:data}

All models are trained on WebFace4M~\cite{Zhu2021WebFace260M}, a
large-scale face dataset containing approximately 4 million images of
205,990 identities.
WebFace4M is well-suited to this study because its size permits
creating training subsets that span several orders of magnitude in
identity count while still providing sufficient images per identity
for meaningful embedding statistics.
Every probe image is aligned to a canonical $112\!\times\!112$ crop with the
standard InsightFace five-landmark pipeline.

\subsection{Ablation Design}
\label{sec:ablation}

We vary four factors in a fully crossed design.

\textbf{Backbone capacity.}
We train three IResNet architectures:
IResNet-34, IResNet-50, and IResNet-100~\cite{he2016deep,deng2019arcface}.

\textbf{Angular-margin loss head.}
We pair each backbone with three angular-margin loss heads:
ArcFace~\cite{deng2019arcface} ($s\!=\!64$, $m\!=\!0.5$),
CosFace~\cite{wang2018cosface} ($s\!=\!64$, $m\!=\!0.4$), and
MagFace~\cite{meng2021magface} ($s\!=\!64$, per-sample $m \in [0.45, 0.8]$ depending on the un-normalised feature magnitude).

\textbf{Training-set size ($\nids$).}
To study the effect of identity coverage, we create five partitions by
randomly sampling $\nids \in \{1\text{K}, 10\text{K}, 50\text{K},
100\text{K}\}$ identities (with a fixed seed), plus one partition that
uses all available identities.
The identities not selected for training serve as same-domain
non-members in the held-out WebFace4M benchmark.

\textbf{Training duration.}
We train the backbones for 5, 10, 20, and 40 epochs to track the 
evolution of cluster-statistic separability.
Longer training exposes each identity to more gradient updates and
may therefore affect the gap between member and non-member statistic
distributions.

The full grid spans
$3~\text{backbones} \times 3~\text{losses} \times 5~\nids\text{ levels}
\times 4~\text{epochs} = 180$ model configurations. All models are trained 
with SGD (momentum 0.9, weight decay $5\!\times\!10^{-4}$), a polynomial-decay 
learning-rate schedule starting at $0.1$ with one warmup epoch, batch size 128, 
mixed-precision (FP16), and random horizontal flips.

\subsection{Evaluation Benchmarks}
\label{sec:benchmarks}

The nine benchmarks (Table~\ref{tab:datasets}) supply standard verification protocols for measuring model quality and provide non-member pools with diverse domain characteristics so that we can study how pose, quality, and demographics affect the apparent cluster-statistic separability.

Eight of the nine benchmarks are drawn from sources disjoint from WebFace4M (we verified that no identity overlap exists by cross-checking subject identifiers); Table~\ref{tab:datasets} lists their sizes and available covariate metadata. Two dataset pairs form natural experiments (matched comparisons where a single attribute varies). \emph{CFP-FF} vs.\ \emph{CFP-FP}~\cite{Sengupta2016CFP} contrasts frontal-only against frontal-plus-profile probes for the same 500 identities, so any statistic difference is attributable to pose. \emph{LFW}~\cite{LFWTech} vs.\ \emph{XQLFW}~\cite{Knoche2021XQLFW} shares the same identity set and pair protocol but degrades one image per pair to isolate image quality. \emph{CPLFW}~\cite{zheng2018cross} (cross-pose LFW), \emph{AgeDB-30}~\cite{Moschoglou2017AgeDB} (age gap $\geq$30\,yr), \emph{RFW}~\cite{Wang2019RFW} (ethnicity-balanced, four groups), and \emph{IJB-C}~\cite{Maze2018IJBC} complete the cross-domain set.

\textbf{Held-out WebFace4M split.}
The ninth benchmark is constructed from WebFace4M itself: for each
model trained on $\nids$ identities, we sample
$\min(\nids,\, 2{,}500)$ members and 2,500 non-members from the
remaining subjects; at $\nids{=}$all no non-members remain.
As both groups share the same acquisition pipeline, AUCs are 
arguably attributable to the cluster-geometry difference between seen 
and unseen identities rather than to domain shift.

\section{Results and Discussion}
\label{sec:results}

This section presents experimental results and analysis for the 180-model grid described in Section~\ref{sec:setup}. We examine verification performance, the effect of training-set size and duration on membership separability, distributional confounds, and statistic fusion. Full per-epoch, per-model results and ROC/DET curves are provided in the supplementary material.

\begin{table}[!t]
\centering
\caption{%
  \textbf{What controls membership information in cluster geometry?}
  \textbf{Top}: partial~$\hat{\eta}^2_p$ from a Type-II ANOVA
  (each factor tested after adjusting for all others) on the AUC
  of a threshold classifier separating member from non-member
  identities ($N\!=\!180$ models);
  values close to 1 indicate that the factor explains nearly all
  variance in this separability.
  \textbf{Bottom}: distributional-shift effect sizes
  (Cohen's~$d'$, Spearman~$\rho$,
  Kruskal--Wallis~$\epsilon^2$) averaged over all models;
  larger absolute values indicate a stronger confound.
}
\label{tab:anova}
\setlength{\tabcolsep}{4pt}
\renewcommand{\arraystretch}{1.05}
\footnotesize
\begin{tabular}{@{} l >{\bfseries}c >{\bfseries}c >{\bfseries}c >{\bfseries}c @{}}
  \toprule
  & \textit{PairCos} & \textit{$\kappa$} & \textit{PenLogit} & \textit{ProtoCE} \\
  \rowcolor{red!6}
  \multicolumn{5}{c}{\normalfont\textbf{Model-level ANOVA} (partial $\hat{\eta}^2_p$)} \\
  $n_\mathrm{ids}$ & \normalfont .99 & \normalfont .95 & \normalfont .99 & \normalfont .95 \\
  Epoch & \normalfont .99 & \normalfont .89 & \normalfont .98 & \normalfont .92 \\
  Backbone & \normalfont .09 & \normalfont .01 & \normalfont .02 & \normalfont .04 \\
  Loss & \normalfont .04 & \normalfont .01 & \normalfont .01 & \normalfont .01 \\
  $n_\mathrm{ids}{\times}$Ep & \normalfont .98 & \normalfont .92 & \normalfont .97 & \normalfont .92 \\
  Residual & \normalfont .00 & \normalfont .03 & \normalfont .01 & \normalfont .02 \\
  \midrule
  \rowcolor{gray!8}
  \multicolumn{5}{c}{\normalfont\textbf{Marginal means} (PairCos AUC)} \\
  \multicolumn{5}{c}{\normalfont $n_\mathrm{ids}$:\;1K\;0.90 $\cdot$ 10K\;0.81 $\cdot$ 50K\;0.73 $\cdot$ 100K\;0.71 $\cdot$ all\;0.71} \\
  \multicolumn{5}{c}{\normalfont Epoch:\;ep5\;0.69 $\cdot$ ep10\;0.75 $\cdot$ ep20\;0.79 $\cdot$ ep40\;0.85} \\
  \midrule
  \rowcolor{green!6}
  \multicolumn{5}{c}{\normalfont\textbf{Distributional shift} (averaged over all 180 models)} \\
  \textit{Cohen's $d'$} & & & & \\
  XQLFW $-$ LFW & \normalfont $-1.64$ & \normalfont $-0.48$ & \normalfont $-1.81$ & \normalfont $+1.13$ \\
  CFP-FP $-$ CFP-FF & \normalfont $-2.27$ & \normalfont $-1.13$ & \normalfont $-2.98$ & \normalfont $+1.14$ \\
  CPLFW $-$ LFW & \normalfont $-1.72$ & \normalfont $-0.72$ & \normalfont $-2.06$ & \normalfont $+0.31$ \\
  \cmidrule(l){1-5}
  \textit{Spearman $\rho$} & & & & \\
  AgeDB (age $\sigma$) & \normalfont $+0.01$ & \normalfont $-0.26$ & \normalfont $+0.17$ & \normalfont $+0.33$ \\
  \cmidrule(l){1-5}
  \textit{Kruskal-Wallis $\epsilon^2$} & & & & \\
  RFW (4 groups) & \normalfont $0.017$ & \normalfont $0.017$ & \normalfont $0.018$ & \normalfont $0.123$ \\
  \bottomrule
\end{tabular}


\end{table}

\subsection{Verification Performance}
\label{sec:verif}

We first establish that our model grid spans a meaningful range of
verification quality.
Table~\ref{tab:eer} reports the verification equal error rate (EER)
at epoch\,40 for every combination of backbone, loss head, and
$\nids$ on the cross-domain benchmarks.
The table is dense and provided primarily as a per-cell reference,
and the discussion below focuses on patterns and points back to
specific cells only as needed.
Performance generally improves with $\nids$, with diminishing
gains beyond $100$K identities.
Once $\nids$ is held fixed, backbone depth provides small reductions in
verification EER on domain-shifted benchmarks, and
the choice of loss head has an even smaller effect.
Notably, the highest cluster-statistic separability occurs at
$\nids{=}1$K, where verification performance is worst.
As $\nids$ grows, verification improves while separability
decreases. However, non-trivial separability persists even at
$\nids$ levels that achieve competitive verification EER, so the
patterns reported below are not confined to poorly performing models.

Table~\ref{tab:eer} also reports the threshold-classifier AUC for
all four cluster statistics.
No single statistic achieves the highest AUC across all benchmarks.
\metric{PenLogit} ranks first most often at moderate-to-high
$\nids$, in particular on the pose- and quality-degraded sets,
whereas \metric{ProtoCE} attains the highest AUC on the frontal
and high-quality sets (CFP-FF, LFW, IJB-C), so the best statistic
remains dataset-dependent.
Because the eight non-WebFace4M benchmarks differ from the training
source in pose, quality, and demographics, the observed AUC
conflates training-attributable and domain-induced differences,
which we discuss in Section~\ref{sec:confounds}. All four statistics 
exhibit similar qualitative trends across factors, so we show only 
\metric{PairCos} distributions in Figure~\ref{fig:main-grid}.

\begin{table*}[t]
\centering
\caption{%
  \textbf{Verification performance and membership separability at epoch\,40.}
  \textbf{E}\,=\,EER,
  \textbf{C}\,=\,PairCos,
  $\boldsymbol{\kappa}$\,=\,vMF-$\kappa$,
  \textbf{L}\,=\,PenLogit,
  \textbf{S}\,=\,ProtoCE AUC.
  \textcolor{green}{Green}\,=\,low EER / high AUC,
  \textcolor{red}{red}\,=\,high EER / low AUC;
  \textcolor{cyan}{cyan}\,=\,$\nids{=}$all AUC~$\geq 0.8$;
  \textbf{bold}\,=\,best per backbone. Lower EER means better recognition, and higher AUC means more membership leakage.
}
\label{tab:eer}
\resizebox{\textwidth}{!}{%
\setlength{\tabcolsep}{3.5pt}
\renewcommand{\arraystretch}{1.15}
\begin{tabular}{@{} c c @{\hspace{6pt}}  |ccccc|ccccc|ccccc|ccccc|ccccc|ccccc|ccccc|ccccc  @{}}
\toprule
& \multicolumn{1}{c @{\hspace{6pt}}}{} &
\multicolumn{5}{c}{\textbf{LFW}} &
\multicolumn{5}{c}{\textbf{XQLFW}} &
\multicolumn{5}{c}{\textbf{CPLFW}} &
\multicolumn{5}{c}{\textbf{RFW}} &
\multicolumn{5}{c}{\textbf{CFP-FP}} &
\multicolumn{5}{c}{\textbf{CFP-FF}} &
\multicolumn{5}{c}{\textbf{AgeDB}} &
\multicolumn{5}{c}{\textbf{IJB-C}} \\
\cmidrule(lr){3-7}\cmidrule(lr){8-12}\cmidrule(lr){13-17}%
\cmidrule(lr){18-22}\cmidrule(lr){23-27}\cmidrule(lr){28-32}%
\cmidrule(lr){33-37}\cmidrule(lr){38-42}
\shortstack{\textsc{Backbone}\\Head} & \multicolumn{1}{c @{\hspace{6pt}}}{} &
1K & 10K & 50K & 100K & All & 1K & 10K & 50K & 100K & All & 1K & 10K & 50K & 100K & All & 1K & 10K & 50K & 100K & All & 1K & 10K & 50K & 100K & All & 1K & 10K & 50K & 100K & All & 1K & 10K & 50K & 100K & All & 1K & 10K & 50K & 100K & All \\
\bottomrule
\end{tabular}}

\vspace{-10pt} 

\end{table*}

\subsection{Effect of $\nids$ and Training Duration}
\label{sec:nids}

\begin{figure*}[t]
  \centering
  \begin{subfigure}[b]{\linewidth}
    \centering
    \includegraphics[width=0.68\linewidth]{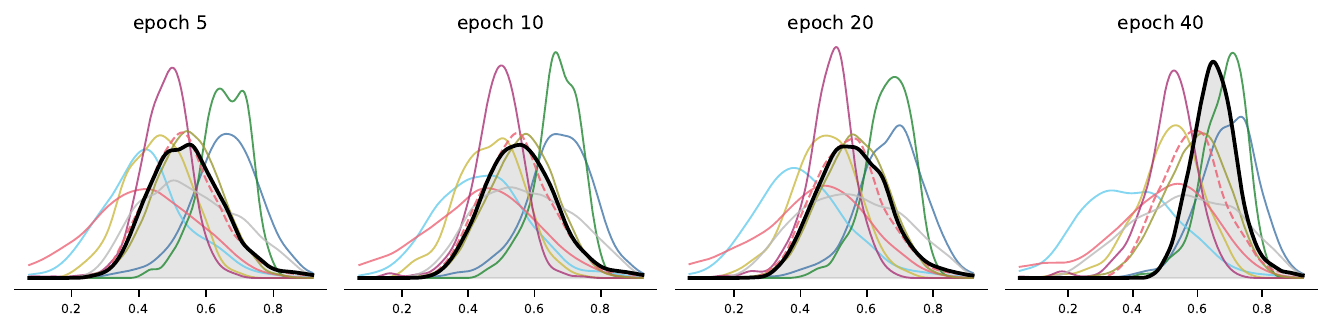}
    \caption{IR-100 / MagFace / 100K identities.}
    \label{fig:grid-a}
  \end{subfigure}\\[2pt]
  \begin{subfigure}[b]{\linewidth}
    \centering
    \includegraphics[width=0.68\linewidth]{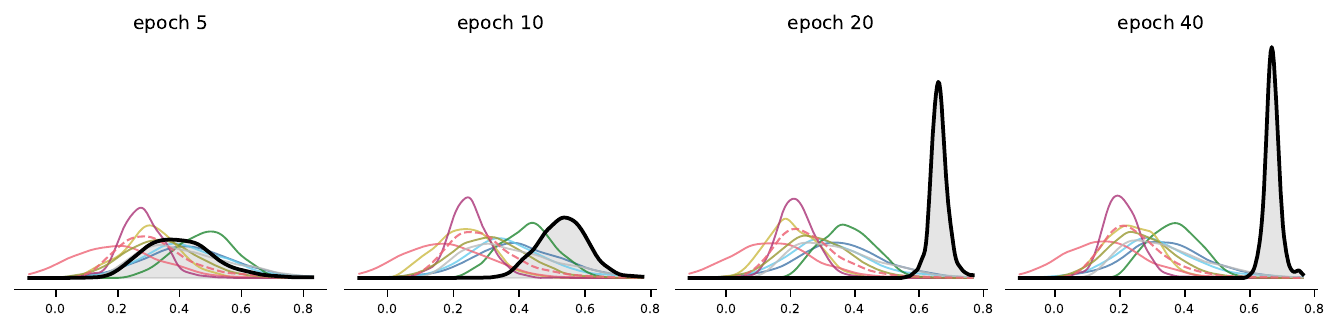}
    \caption{IR-100 / MagFace / 1K identities.}
    \label{fig:grid-b}
  \end{subfigure}\\[2pt]
  \begin{subfigure}[b]{\linewidth}
    \centering
    \includegraphics[width=0.85\linewidth]{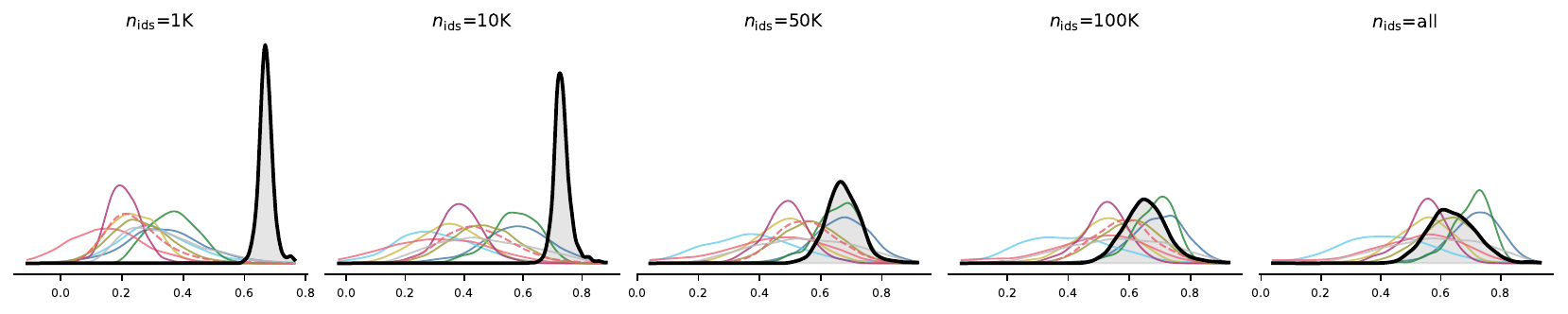}
    \caption{IR-100 / MagFace / epoch\,40.}
    \label{fig:grid-c}
  \end{subfigure}\\[4pt]
  %
  %
  \includegraphics[width=\linewidth]{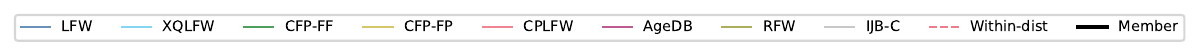}
  \caption{%
    \metric{PairCos} ($\bar{s}$) kernel density estimates.
    The bold black curve (with shading) shows \emph{member}
    identities (present in the training set).
    All other curves show \emph{non-member} identities:
    the nine thin solid lines correspond to cross-distribution
    benchmarks whose domain differs from the training data
    (one colour per dataset; see legend),
    while the dashed line (``Within-dist'') shows non-members from
    the held-out WebFace4M split, which shares the training domain
    (no domain shift).
    (a,\,b)~vary training epochs at fixed $\nids$;
    (c)~varies $\nids$ at epoch\,40.
  }
  \label{fig:main-grid}
  \vspace{-1.5em}
\end{figure*}

Figure~\ref{fig:main-grid} presents the main finding of this study.
Panel~(c) shows that the \metric{PairCos} distributions of members
and non-members converge as $\nids$ increases from 1K to all
identities. At $\nids{=}1$K the distributions are well separated,
whereas at $\nids{=}100$K and beyond the separation narrows,
though it remains above chance even on the held-out WebFace4M split
(Table~\ref{tab:mlp}, discussed in Section~\ref{sec:fusion}).

Panels~(a) and~(b) of Figure~\ref{fig:main-grid} isolate the effect
of training duration at two extremes of identity coverage.
At $\nids{=}1$K, the \metric{PairCos} separation between
members and non-members increases with the number of training epochs.
At $\nids{=}100$K the same trend is present but attenuated.

Table~\ref{tab:anova} reports the ANOVA results
($\hat{\eta}^2_p$, the proportion of
variance explained after accounting for all other factors).
Training-set size explains the most variance
($\hat{\eta}^2_p \geq .95$), followed by training duration
($\hat{\eta}^2_p \geq .89$).
The strong $\nids\!\times\!\text{Epoch}$ interaction
($\hat{\eta}^2_p \geq .92$) further shows that the epoch effect
is moderated by identity coverage.
The $\nids\!\times\!\text{Epoch}$ cell means on the
same-domain WebFace4M split show that the mean \metric{PairCos}
AUC increase from epoch\,5 to epoch\,40 ranges from
$+.38$ at $\nids{=}10$K down to $+.16$ at $\nids{=}100$K,
with $\nids{=}1$K reaching a ceiling near 1.0 already at
epoch\,10.
Models with small identity pools thus show the largest separability
gains over training, whereas models trained on $100$K identities
have only a modest increase.

\textbf{Backbone and loss head.}
The ANOVA also shows that backbone architecture and loss head
explain far less variance in threshold-classifier AUC, the
loss-head factor reaching at most $\hat{\eta}^2_p = .04$ and
backbone depth at most $.09$ across all four statistics --- an order
of magnitude below the contributions of $\nids$ and
number of epochs.
Beyond the IResNet grid, a ViT-T~\cite{dosovitskiy2021vit,touvron2021deit}
backbone (CosFace) trained under the identical protocol~\cite{an2022partialfc}
reproduces the same $\nids$ trend in both EER and membership AUC
(Table~\ref{tab:eer}, bottom block), so the effect is not specific to
convolutional backbones.

\subsection{Effect of Input-Data Distribution}
\label{sec:confounds}

The benchmarks used in this study differ in pose, quality, age, and
demographics, which affect embedding-cluster geometry independently
of membership.
The distributional-shift rows of Table~\ref{tab:anova} quantify
these effects, and we discuss each factor below.
For matched-pair benchmarks, we also translate each effect into
the AUC change it would produce at the same-domain
training-attributable baseline at $\nids{=}100$K (where the AUC
is $\approx 0.71$).
We do so by passing each AUC through the inverse standard normal
cumulative distribution function (a probit transform), on which
equal shifts in the underlying non-member distribution correspond
to equal numerical changes regardless of baseline.
We take the matched-pair difference on this transformed scale
and convert it back to AUC at the chosen baseline.
This step is needed because the AUC scale saturates near 1, so a
fixed shift in the underlying distribution produces a smaller AUC
change near the ceiling than in the middle.

\textbf{Pose.}
Profile-view faces produce large shifts in cluster-statistic
distributions.
The CFP-FP vs.\ CFP-FF comparison is the cleanest controlled
experiment: both sets share the same 500 identities, so the
observed Cohen's $d'$ (standardised mean difference) is
attributable to pose alone (Table~\ref{tab:anova}).
CPLFW vs.\ LFW yields even larger $|d'|$ for \metric{PairCos}
and \metric{PenLogit}, but those two sets are not identity-matched,
so other factors may contribute.
In both comparisons, profile views widen the
intra-identity distribution regardless of
membership.
This pushes statistic values towards the non-member range
and inflates separability.
At the $\nids{=}100$K baseline, the pose contribution from
CFP-FF to CFP-FP corresponds to an AUC change of $+0.26$,
matching or exceeding the training contribution above
chance ($+0.21$).

\textbf{Image quality.}
XQLFW shares LFW's identity set and pair protocol but
degrades one image per pair. For \metric{PairCos} and
\metric{PenLogit}, the resulting Cohen's $d'$ is of the same
order as that of cross-pose variation though somewhat smaller;
$\kappa$ shows a smaller quality effect ($|d'|{=}0.48$ vs.\
$1.13$ for pose).
Lower image quality increases intra-identity spread in embedding
space and shifts cluster-statistic distributions in the same direction
as pose variation.
At the same baseline, the quality contribution from LFW to XQLFW
corresponds to an AUC change of $+0.22$, comparable to the
pose contribution.

\textbf{Age.}
Within AgeDB-30, the Spearman rank correlation $\rho$
(a non-parametric measure of monotonic association)
between within-identity age spread and each cluster statistic
(Table~\ref{tab:anova}) is weak and varies in sign across
statistics.
Only $\kappa$ exhibits the expected negative association
($\rho = -0.26$), under which identities spanning wider age
ranges form less concentrated clusters, whereas \metric{PairCos}
is essentially uncorrelated ($\rho = +0.01$) and both
\metric{PenLogit} and \metric{ProtoCE} carry small positive
correlations ($\rho = +0.17$ and $\rho = +0.33$).
Age-range variation therefore exerts the weakest and least
consistent influence on cluster geometry among the confounds
examined here.

\textbf{Ethnicity.}
A Kruskal--Wallis test (non-parametric one-way ANOVA) across the
four RFW demographic groups
(African, Asian, Caucasian, Indian) yields
$\epsilon^2 \approx 0.02$ (small) for \metric{PairCos}, $\kappa$,
and \metric{PenLogit},
and $\epsilon^2 = 0.12$ (medium) for \metric{ProtoCE}.
Cluster-statistic values therefore vary across demographic groups
within the non-member pool, with \metric{ProtoCE} again being
the most sensitive statistic.
Although these effect sizes are modest relative to the
$\nids$ factor, they indicate that the demographic composition
of a non-member benchmark affects the baseline against which
membership is measured.

Pose and quality produce the largest shifts in cluster-statistic
distributions, with demographic composition and age contributing
smaller and, for age, less consistent effects.
The matched-pair pose and quality contributions show that the
domain-induced component can match or exceed the training
contribution itself at higher $\nids$, so the threshold-classifier
AUC on any non-WebFace4M benchmark cannot be interpreted as
membership signal alone.

\subsection{Statistic Fusion}
\label{sec:fusion}

To test whether combining statistics recovers more membership
information, we train a two-hidden-layer MLP
($64{\to}32$ units, ReLU, BatchNorm, dropout\,0.3;
Adam with $\eta=10^{-3}$)
on all four cluster statistics.
For each epoch-40 configuration we perform five-fold stratified
cross-validation with an 80/20 identity split on same-domain
WebFace4M identities, reserving 15\% of each training fold as
a validation set for early stopping.
\begin{table}[!t]
\centering
\caption{%
  \textbf{Statistic fusion recovers more membership information than
  any single statistic.}
  Same-domain held-out split (epoch\,40, 5-fold stratified CV).
  Each cell shows best single-statistic AUC\,/\,fold-averaged MLP AUC.
  \textbf{Bold} highlights the MLP gain over the best single statistic.
}
\label{tab:mlp}
\setlength{\tabcolsep}{3pt}
\renewcommand{\arraystretch}{1.05}
\footnotesize
\scalebox{0.90}{
\begin{tabular}{@{} ll cccc @{}}
  \toprule
  & & \multicolumn{4}{c}{$\nids$} \\
  \cmidrule(l){3-6}
  Backbone & Loss & 1K & 10K & 50K & 100K \\
  \midrule
  \multirow{3}{*}{IR-34}
    & Arc & 1.00\,/\,\textbf{1.00} & .98\,/\,\textbf{1.00} & .80\,/\,\textbf{.87} & .65\,/\,\textbf{.69} \\
    & Cos & .99\,/\,\textbf{1.00} & .98\,/\,\textbf{.99} & .79\,/\,\textbf{.88} & .65\,/\,\textbf{.71} \\
    & Mag & 1.00\,/\,\textbf{1.00} & .98\,/\,\textbf{1.00} & .79\,/\,\textbf{.85} & .64\,/\,\textbf{.68} \\
  \cmidrule(l){1-6}
  \multirow{3}{*}{IR-50}
    & Arc & 1.00\,/\,\textbf{1.00} & .99\,/\,\textbf{1.00} & .84\,/\,\textbf{.91} & .68\,/\,\textbf{.73} \\
    & Cos & .99\,/\,\textbf{1.00} & .98\,/\,\textbf{.99} & .82\,/\,\textbf{.91} & .68\,/\,\textbf{.75} \\
    & Mag & 1.00\,/\,\textbf{1.00} & .98\,/\,\textbf{1.00} & .82\,/\,\textbf{.89} & .67\,/\,\textbf{.72} \\
  \cmidrule(l){1-6}
  \multirow{3}{*}{IR-100}
    & Arc & 1.00\,/\,\textbf{1.00} & .99\,/\,\textbf{1.00} & .86\,/\,\textbf{.92} & .73\,/\,\textbf{.79} \\
    & Cos & .99\,/\,\textbf{1.00} & .98\,/\,\textbf{.99} & .84\,/\,\textbf{.92} & .71\,/\,\textbf{.80} \\
    & Mag & 1.00\,/\,\textbf{1.00} & .98\,/\,\textbf{1.00} & .85\,/\,\textbf{.91} & .72\,/\,\textbf{.79} \\
  \bottomrule
\end{tabular}
}

\vspace{-14pt} 

\end{table}

Table~\ref{tab:mlp} reports the best single-statistic AUC alongside
the fold-averaged MLP AUC.
The MLP consistently matches or outperforms the best individual statistic,
with the largest gains at $\nids{=}50\text{K}$ and $100\text{K}$
where individual statistics are weakest
(one-sided Wilcoxon signed-rank: $W{=}666$, $p<10^{-10}$).

\subsection{Limitations}
\label{sec:limitations}

\textbf{Generalisation across training data.}
All experiments use a single training source (WebFace4M).
Whether the trends generalise to other training datasets is an open
question, as comparably large datasets are not publicly available or have been retracted.

\textbf{Experimental controls.}
We do not evaluate defences such as
DP-SGD~\cite{Abadi2016DP} or knowledge
distillation~\cite{Hinton2015Distillation}.
The number of probe images~$k$ per identity varies across
benchmarks and is not controlled for, which may affect estimates
of the cluster statistics non-uniformly.
The held-out split is constructed by random partitioning;
in practice, membership boundaries may correlate with
identity-level attributes.

\textbf{Statistical assumptions and scope.}
Because our four statistics capture only cluster geometry, the
measured separability is a lower bound on the membership
information available from embeddings; a richer model operating
on raw embedding vectors could extract more.
The probit-scale translation in Section~\ref{sec:confounds} assumes
approximately Gaussian member and non-member statistic
distributions and stable variances across matched pairs.

\section{Conclusion and Future Work}
\label{sec:conclusion}

We evaluated how much membership information is retained in the
embedding geometry of 180 open-set face recognition models across
nine benchmarks.
The results lead to three practical conclusions.
First, training-set size has the largest effect on member/non-member
separability, while backbone architecture and loss head contribute
far less.
Training on more identities is therefore the only design choice we
tested that substantially reduces the geometric membership signal.
Second, cross-domain non-member benchmarks inflate
the apparent signal by conflating domain shift with
training-attributable differences.
Privacy audits should therefore use same-domain held-out references,
since cross-domain non-members overestimate the apparent membership signal.
Third, even at high $\nids$ where individual statistics are weak,
fusing all statistics with a learned classifier recovers
additional membership information (Table~\ref{tab:mlp}). As our statistics 
capture only cluster geometry, the measured
separability is a lower bound on the membership information
available from embeddings.
Future work includes extending these benchmarks to other backbones 
and evaluating whether defences such as~\cite{Abadi2016DP} can 
reduce the geometric signal without degrading performance.

\section*{Acknowledgments}
This work has received funding from the European Union's Horizon
Europe research and innovation programme under Grant Agreement
No.~101189650 (CERTAIN: \emph{Certification for Ethical and Regulatory
Transparency in Artificial Intelligence}), and the Swiss State Secretariat
for Education, Research and Innovation (SERI).

{\small
\bibliographystyle{ieee}
\bibliography{references}
}

\end{document}